\documentclass{article}

\usepackage[numbers]{natbib}
\usepackage[final]{ap_2018}
\usepackage{ap_2018}

\usepackage[utf8]{inputenc} 
\usepackage[T1]{fontenc}    
\usepackage{hyperref}       
\usepackage{url}            
\usepackage{booktabs}       
\usepackage{amsfonts}       
\usepackage{nicefrac}       
\usepackage{microtype}      
\usepackage{footnote}
\usepackage{graphics}
\usepackage{graphicx}
\makesavenoteenv{tabular}
\makesavenoteenv{table}
\usepackage{amsmath}
\usepackage{microtype}
\usepackage{array}
\usepackage[noend]{algorithm2e}
\usepackage{wrapfig}
\usepackage{color}
\usepackage{floatrow}
\usepackage{blindtext}
\usepackage{subcaption}
\usepackage{capt-of}

\newfloatcommand{capbtabbox}{table}[][\FBwidth]

\title{SlugBot: Developing a Computational Model and Framework of a Novel Dialogue Genre}

\author{
	Kevin K. Bowden \and Jiaqi Wu \and Wen Cui \and Juraj Juraska \and Vrindavan Harrison \and Brian Schwarzmann \and Nick Santer \and Marilyn Walker\\
	Natural Language and Dialogue Systems Laboratory\\
	University of California Santa Cruz\\
	\texttt{\{kkbowden, jwu64, wcui7, jjuraska, vharriso, brschwar, nsanter, mawalker\}@ucsc.edu} \\
}

\begin{document}

\maketitle

\begin{abstract}

One of the most interesting aspects of the Amazon Alexa Prize
competition is that the framing of the competition requires the
development of new computational models of dialogue and its structure.
Traditional computational models of dialogue are of two types: (1)
task-oriented dialogue, supported by AI planning models, or simplified
planning models consisting of frames with slots to be filled; or (2)
search-oriented dialogue where every user turn is treated as a search
query that may elaborate and extend current search results.  Alexa
Prize dialogue systems such as SlugBot must support conversational
capabilities that go beyond what these traditional models can
do. Moreover, while traditional dialogue systems rely on theoretical
computational models, there are no existing computational theories
that circumscribe the expected system and user behaviors in the
intended conversational genre of the Alexa Prize Bots.  This paper
describes how UCSC's SlugBot team has combined the development of a
novel computational theoretical model, Discourse Relation Dialogue
Model, with its implementation in a modular system in order to test
and refine it. We highlight how our novel dialogue model has led us to
create a novel ontological resource, UniSlug, and how the structure of
UniSlug determines how we curate and structure content so that our
dialogue manager implements and tests our novel computational dialogue
model.
\end{abstract}

\section{Introduction}
\label{sec:intro}

One of the most interesting aspects of the Amazon Alexa Prize
competition is that the framing of the competition requires the
development of new computational models of dialogue and its structure.
Traditional computational models of dialogue are of two types: (1)
task-oriented dialogue, supported by AI planning models or simplified
planning models consisting of frames with slots to be filled 
  \cite{GS86,Litman85,AP80,shah2018bootstrapping,Walkeretal97}; or (2)
search-oriented dialogue where every user turn is treated as a new
query that either starts a new dialogue segment or extends current
search results.  These traditional models and existing systems
that are built using them are based on several simplifying
assumptions:

\begin{itemize}
\item {\sc Search Model}: Conversations are simply sequences of user search queries (initiated by the user) and search results \cite{Duplessisetal16,Charrasetal16,krause2017redundancy};
\item {\sc Task Model}: Conversations are composed of sequences of specific tasks such
as setting a timer, booking a flight, shopping for specific items
  \cite{GS86,Litman85,AP80,shah2016interactive};
\item  {\sc Script Model}: Conversations follow a finite-state script and
scripts can be written by hand to support all the conversations and
conversational variants that a user might want to have with the system
\cite{bellegarda2013large,guzzoni2007active,kamm1997evaluating}.
\end{itemize}

However, the Alexa prize requires the development of {\bf an
  open-domain conversational agent that can talk about any topic and
  carry on a conversation for at least 20 minutes}. This framing means
that the dialogue genre of Alexa Prize systems requires not only
system development, but the development of new computational and
theoretical models of dialogue.  It is also important to note that
although Alexa Prize bots are characterized as open domain bots, much
of the recent work on open domain dialogue has been focused on chat,
with attempts for example to train systems using the "Open
Subtitles" corpus \cite{Nio2014, LisonTK18opensubtitles}.  In contrast, SlugBot must
utilize substantive and up-to-date content on news, movies, books,
fashion, technology, news entities, actors, and other topics. This
content must be scraped daily from relevant sources and structured in
a way that makes it possible to support a coherent conversation
about any of these topics. Thus content creation, curation, and
structuring is a substantive task all by itself.

Here, in addition to describing the
SlugBot system, we describe elements of a new dialogue model which we
call {\sc discourse relation dialogue model} ({\bf DRDM}). The features of
our proposed DRDM are:
\begin{itemize}
\item Mixed Initiative: Novel dialogue strategies are needed that will allow
  SlugBot to take the initiative in conversation. It will not be
  possible to carry on a 20 minute conversation if SlugBot is simply
  responding to user initiatives as is assumed by the {\sc search
    model} above.

\item Discourse Relations: Novel models of discourse coherence in
  dialogue are needed. We propose a model based on a framework of Penn
  Discourse TreeBank discourse relations
  \cite{HE79,Hobbs78,Louisetal10,prasad-etal-lrec08}. Discourse
  coherence in {\sc task model} dialogue systems arises from the
  structure of the task.  Discourse coherence in {\sc search model}
  dialogue systems, such as it exists, are driven by the user's search
  intentions. Discourse coherence in {\sc script model} dialogue
  systems is created by the user interaction designer rather than
  being an instantiation of an underlying theory of discourse
  coherence.

\item Knowledge Graph: A large ontology with specific world-knowledge
  can provide DRDM with dialogue strategies based on general,
  re-usable relations between conversational turns. We describe how we
  have developed UniSlug, an ontology based on integrating the schemas
  of several existing ontologies, which we use to drive dialogue
  strategy selection as well as natural language understanding
  \cite{hakkani2014probabilistic}.

\end{itemize}

The fact that there are no existing theoretical or computational
models for the Alexa Prize dialogue genre cannot be overstated. Novel
methods are needed to drive the system behaviors while ensuring
discourse coherence. These methods should be {\bf general} so that
they can be systematically applied to subconversations on {\bf different
  topics} or {\bf user activities}.  Task-oriented dialogue models
assume that coherence arises because both the system and user can
recognize the intentions of their conversational partner as
contributing to the completion of the task, or as meta-dialogue
related to organizing contributions to the completion of the task
\cite{GS86,Litman85,AP80}.  While some dialogue segments in SlugBot
can be modeled as tasks, the overall dialogue structure is not
task-related.  Previous theoretical work on conversation merely
describes aspects of conversational structure without considering
algorithms or models that can drive the behavior of a conversational
agent \cite{gilmartin17,HE79}. We will describe below how we build on
previous models of discourse coherence, but it is also important to
note that models of discourse coherence using discourse relations have
mainly been applied to highly structured texts such as newswire or
student essays
\cite{carlsonMarcuOkurowski01,marcu96,mann88,Prasadetal10,MarcuEchihabi02}.
There is no large annotated corpus showing how discourse relation
models could be applied to dialogue: previous projects in this vein
are merely exploratory or focused around a few examples
\cite{Tonellietal10,asher-lascarides-book03,HE79}.

We will describe in more detail in Section~\ref{sec:system_arch} how
we have combined the development of a novel computational theoretical
model, DRDM, with its implementation in a modular system in order to
test and further develop the model. We highlight our extended efforts
at content curation and content structuring and describe how we have
built new knowledge bases and a novel dialogue manager to support our
novel computational dialogue model.

\section{System Design and Architecture}
\label{sec:system_arch}

The architecture of SlugBot is driven by the need to support DRDM, our
novel computational model of open-domain dialogue. The DRDM dialogue
model relies on UniSlug, a new large domain ontology that we have
built and integrated into SlugBot. The DRDM model is based on two
ideas. First, we 
claim that it is possible to model the coherence of open-domain
dialogue using discourse relations, specifically we currently utilize
the four high level discourse relations used in the Penn Discourse
TreeBank (PDTB), which provide good generalization capabilities and
are compatible with other discourse relation frameworks. These are
framed in terms of relations between abstract objects (as realized by
sentences or clauses). Here we apply these to relations between
utterances across the agent and the user in discourse. 

The top level Penn Discourse TreeBank (PDTB) discourse relations that we utilize are~\cite{Prasadetal10}:
\begin{itemize}
\item {\sc expansion}: The expansion class covers those relations
  which expand the discourse and move its narrative or exposition
  forward. Its subclasses include instantiating a set, restating,
  describe alternative situations and more. This is the most general
and weakest discourse relation, since it covers both continuing
to talk about the same thing, as well as talking about a more specific
attribute of an entity.
\item {\sc comparison}: The comparison class applies when a discourse
  relation is established between Arg1 and Arg2 in order to highlight
  prominent differences between the two situations. Its subtypes are
  {\sc contrast} and {\sc concession}. Disagreements can be viewed
as a type of contrast. 
\item {\sc contingency}: indicates that one of the situations
described in Arg1 and Arg2 causally influences the other.
For example, SlugBot may offer an opinion, along
with the reasons underlying it. Opinions
are causally related to (justified by) these reasons.
\item {\sc temporal}: this relation applies when the situations
  described in the arguments are
  related temporally, either in overlap or in a temporal sequence.
The main use of the temporal relation in SlugBot is in the context
of telling stories, where we take advantage of the
fact that story events are told in temporal sequence.
\end{itemize}

The second idea is that these discourse relations, along with dialogue
acts and named-entities, can be used to guide retrieval of utterances
from the many different sources of content. Other existing retrieval
based chatbots also operate on large existing corpora such as
Twitter~\cite{Nio2014, Higashinaka14opendomain}, the Open Subtitles corpus~\cite{DuplessisLLR16}, or movie scripts \cite{Banchs2012iris, Ameixa2014outdomain},
but none of them use either discourse relations or dialogue acts.
Instead, the criteria by which utterances are retrieved has been based
on their similarity to the current system utterance or a previously
existing reply to the current system utterance. Similarity measures
have been adopted from information retrieval, e.g. they include both
TF-IDF and word-embeddings.

Instead our retrieval mechanism, as described in more detail in
Section~\ref{sec:retrieval-sec} is controlled by a combination of 
dialogue acts, discourse relations and named-entity matching.
The discourse relations in DRDM can be instantiated by different types
of dialogue acts, such as questions or statements as shown in
Table~\ref{table:drmd_ex}, thus the dialogue act that instantiates the
discourse relation must be specified in the dialogue flow.
Table~\ref{table:drmd_ex} illustrates how these relations can be
instantiated in the DRDM. The RHS column provides a dialogue act and
discourse relation specification, as well as an informal description
on how the discourse relation instantiation relies on UniSlug, our new
mondo-domain ontology.

\begin{table}[th]
\begin{small}
\begin{tabular}
{p{0.9cm}|p{7.1cm}|p{4cm}}
\toprule
Agent & {Are you planning on going on any trips soon?}. & Dialogue Act: {\sc yes-no question} \\ \hline 
User & {yes, we are planning to go to Paris} & system recognizes city-name in GeoNames ontology\\  \hline 
Agent & {Ah Paris, What will you do there?} & Dialogue Act: {\sc open question }\\  \hline 
User & {I am going to see the Eiffel Tower} & system recognizes point-of-interest, go up ontology \\  \hline 
Agent & {Are you also going to check out the Louvre} & Dialogue Act: {\sc yes-no question}, {\sc comparison} relation is instantiated by finding a different point of interest for Paris by going down POI ontology \\  \hline 
User & {I don't know} & \\  \hline 
Agent & {Well, I would think about it. The Mona Lisa is there and that could be worth seeing.} & Dialogue Act: {\sc provide-opinion}, {\sc expansion} relation is achieved by following artworks relation on Louvre \\  \hline 
User & {I see. I'll think about it.} & \\  \hline 
Agent & {Are you a fan of any other artwork by Leonardo da Vinci?} & Dialogue Act: {\sc yes-no question}, {\sc comparison} relation is achieved by following artist relation for the Mona Lisa\\  
\bottomrule               
 \end{tabular}
\end{small}
\caption{Expansion and Comparison Discourse Relations using GeoNames Ontology}\label{table:drmd_ex}
\end{table}

\begin{figure}[htb]
  \includegraphics[width=3.5in]{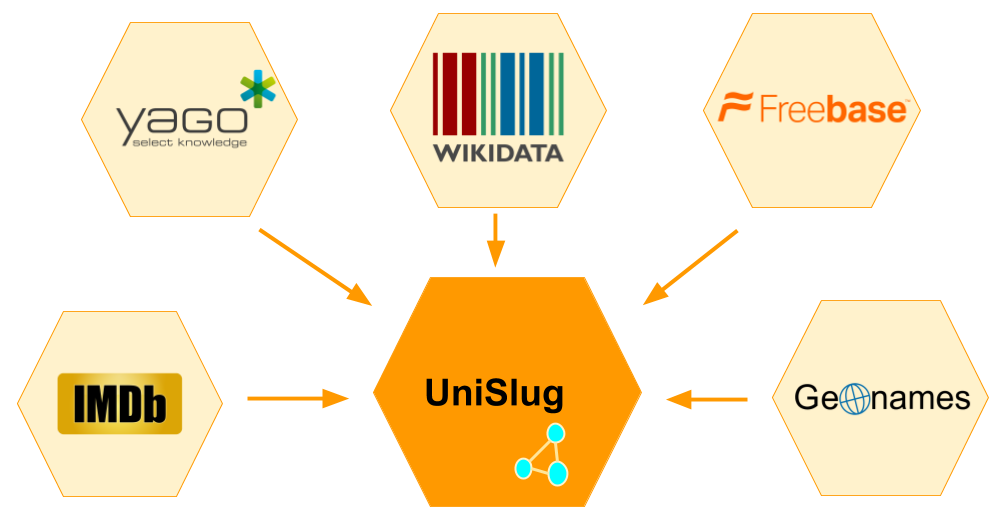}
  \centering
  \caption{UniSlug is SlugBot's large scale integrated ontology
for Dialogue Management and Natural Language Understanding.}
  \label{fig:unislug}
\end{figure}

Figure~\ref{fig:unislug} illustrates the UniSlug ontology that
we have built for SlugBot 
by integrating several existing ontologies 
using the schema from schema.org. We  then use this ontology for
both understanding and dialogue management. The main way that the ontology
is used for dialogue management is illustrated
in Table~\ref{table:drmd_ex}.

\subsection{Architecture}

\begin{figure}[ht]
  \includegraphics[scale=.24]
{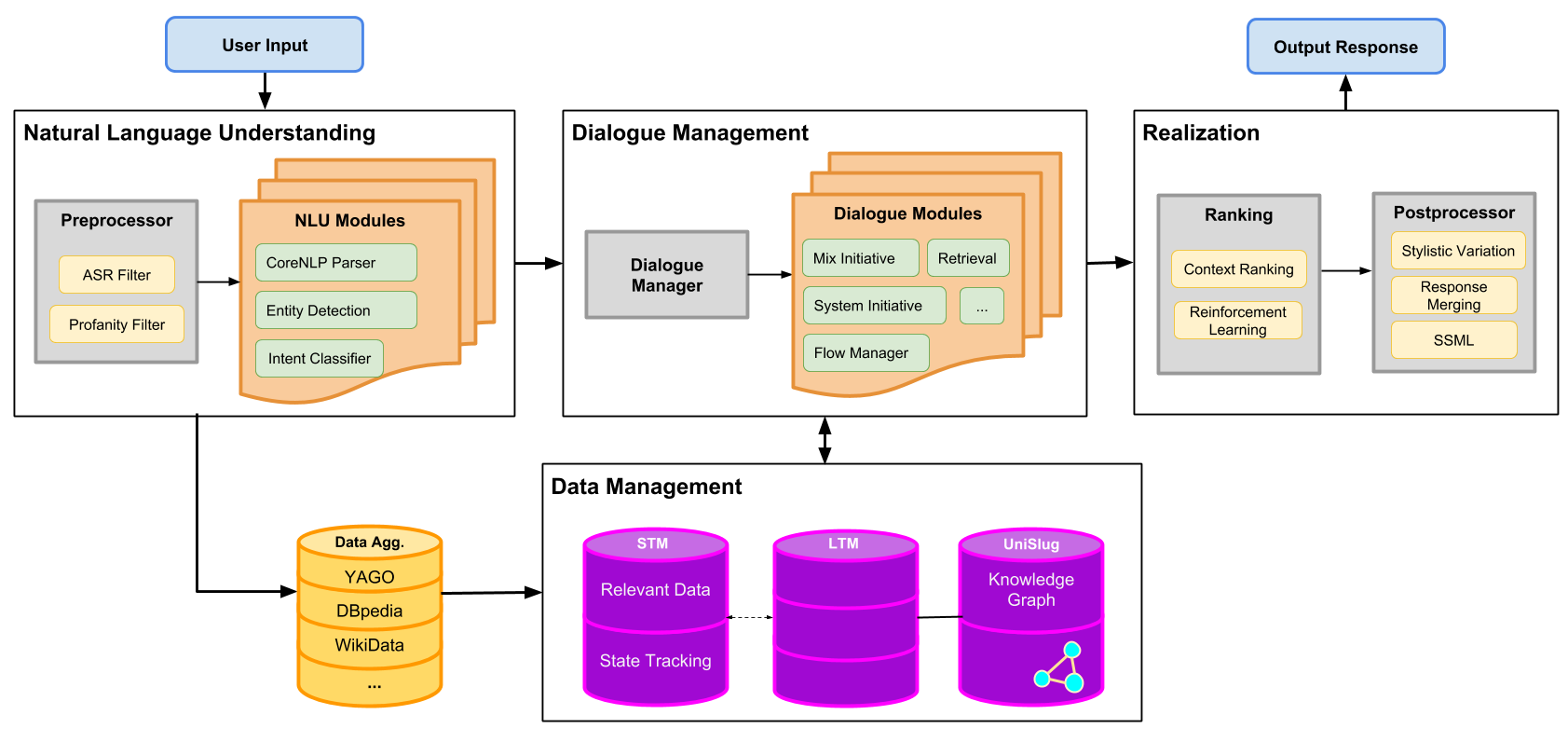}
  \centering
  \caption{SlugBot system architecture.}
  \label{fig:system_arch}
\end{figure}

Figure \ref{fig:system_arch} describes the complete system
architecture and Sections \ref{sec:nlu} - \ref{sec:postprocess}
describe the individual components such as natural language
understanding, dialogue management, data management, and realization
components. In Section \ref{sec:amazon_infrastructure} we will
specifically examine our Amazon Skill infrastructure as shown in
Figure \ref{fig:webapp_arch}.

\subsection{Natural Language Understanding (NLU)}
\label{sec:nlu}
\subsubsection{Preprocessor}
\textbf{ASR Filter:} Our initial preprocessing stage involves analyzing the tokens interpreted by the Automatic Speech Recognition software (ASR). We calculate the average hypothesis scores of the users input and prompt the user for clarification if the score is too low. Intuitively, it is better to ask for clarification rather than misinterpret the user input; however, if we consistently get a low ASR score, we are forced to estimate their utterance. To account for this, we retain all possible ASR interpretations such that we are able to better process the noise in their input.

\textbf{Profanity Filter:} Our profanity filter looks for keywords that typically indicate profanity or possible offensive language. This filter has expanded over time as we became aware of offensive content in Wikipedia, which we thought was a pretty benign source of information but in fact contains references to inappropriate content. In future work, we wish to use our profanity filter to indicate among other things user frustration. Currently, however, we just use the profanity filter to eliminate inappropriate content from the candidate response pool.

\subsubsection{NLU Modules}
\textbf{CoreNLP Parser:} After preprocessing the data we use our Natural Language Understanding
(NLU) engine to create a deep structure representation of the user's
utterance. Our first layer of NLU relies on the Stanford CoreNLP
Toolkit~\cite{stanford14}. Our internal representation is based on the dependency
parse of the respective utterance which is consolidated into a concise
tree using the dependency relations. The part-of-speech (POS) and
sentiment score from CoreNLP are also encoded into this structure. We
do coreference resolution by mapping the coreference tags returned by
CoreNLP to the data stored within our system. 

\textbf{Entity Detection:} We have also developed our own named entity recognizer SlugNERDS \cite{bowden2018slugNERDS, bowden2018opendomain} because the existing named entity recognizers
were not fine-grained enough to support dialogue
interaction. SlugNERDS is based on the Google Knowledge Graph\footnote{\url{https://developers.google.com/knowledge-graph/}} and can take
advantage of the fact that the Google Knowledge Graph is both robust and constantly
updated, allowing us to consistently detect the newest and most obscure named entities. 

\textbf{Intent Classifier:}
Both our NLU and our indexed retrieval mechanism rely on a dialogue
act classification. We develop an utterance intent ontology and
develop a Neural Network model to recognize user intents. The intent
ontology consists of 33 discrete intents. Some example utterances with
their associated intents can be seen in Table
\ref{table:utterence-intents}. The ontology is designed to allow each
intent to be recognized without conversational context. In other words,
it is assumed that each utterance's intent can be determined
irrespective of where they occur in a conversation.

\begin{table}[ht]
\begin{small}
\begin{tabular}{l|l}
\toprule
\bf Intent & \bf Utterance \\ \midrule
\texttt{request\_opinion} & did you like beauty and the beast \\
\texttt{request\_change\_topic} &	no can we do something else \\
\texttt{request\_opinion\_justify}	& why do you like wine \\
\texttt{assertion\_on\_bot} &	you are so much better than siri \\
\texttt{request\_exit}	& can we stop talking right now \\
\texttt{request\_service} & play country christmas songs \\
\texttt{request\_discuss\_topic}	& do you know anything about pizza \\
\texttt{request\_confirm\_understanding}	& are you understanding me \\
\bottomrule
\end{tabular}
\caption{Example utterances with their corresponding intent labels.}
\label{table:utterence-intents}
\end{small}
\end{table}

We use a subset of the CAPC dataset to train the intent classification
model. Utterances were selected from the CAPC dataset and then
annotated to provide input to a supervised learner. This resulted in a
dataset of roughly 50K annotated utterances which is then separated
into training (80\%), development (10\%), and test (10\%) subsets. Next,
we train an intent classifier on the training set, tune the
hyperparameters on the development set, and use the test set for final
evaluation of the model. Our intent classifier is a Neural Network
model that uses a combination of Recurrent Neural Network and
Convolutional Neural Network architectures.

\textbf{Additional NLU:}
Additional NLU components in our pipeline include the NPS dialogue act classifier\cite{NPS}. We further do noun and entity disambiguation to know that, for example a lion is an animal and \textit{Watchmen} is a 2009 American superhero movie. Our topic classification takes into account a broad classifier which is trained using topic annotated news articles in addition to the provided cobot topic classifier. We additionally take into account a more refined topic classifier which directly maps to the 42 topics supported by the Flow Manager described in section \ref{sec:flowman}. 

\subsection{Data Management and UniSlug}
\label{sec:data}
\label{sec:unislug}

\begin{figure}[ht]
  \includegraphics[width=4.0in]{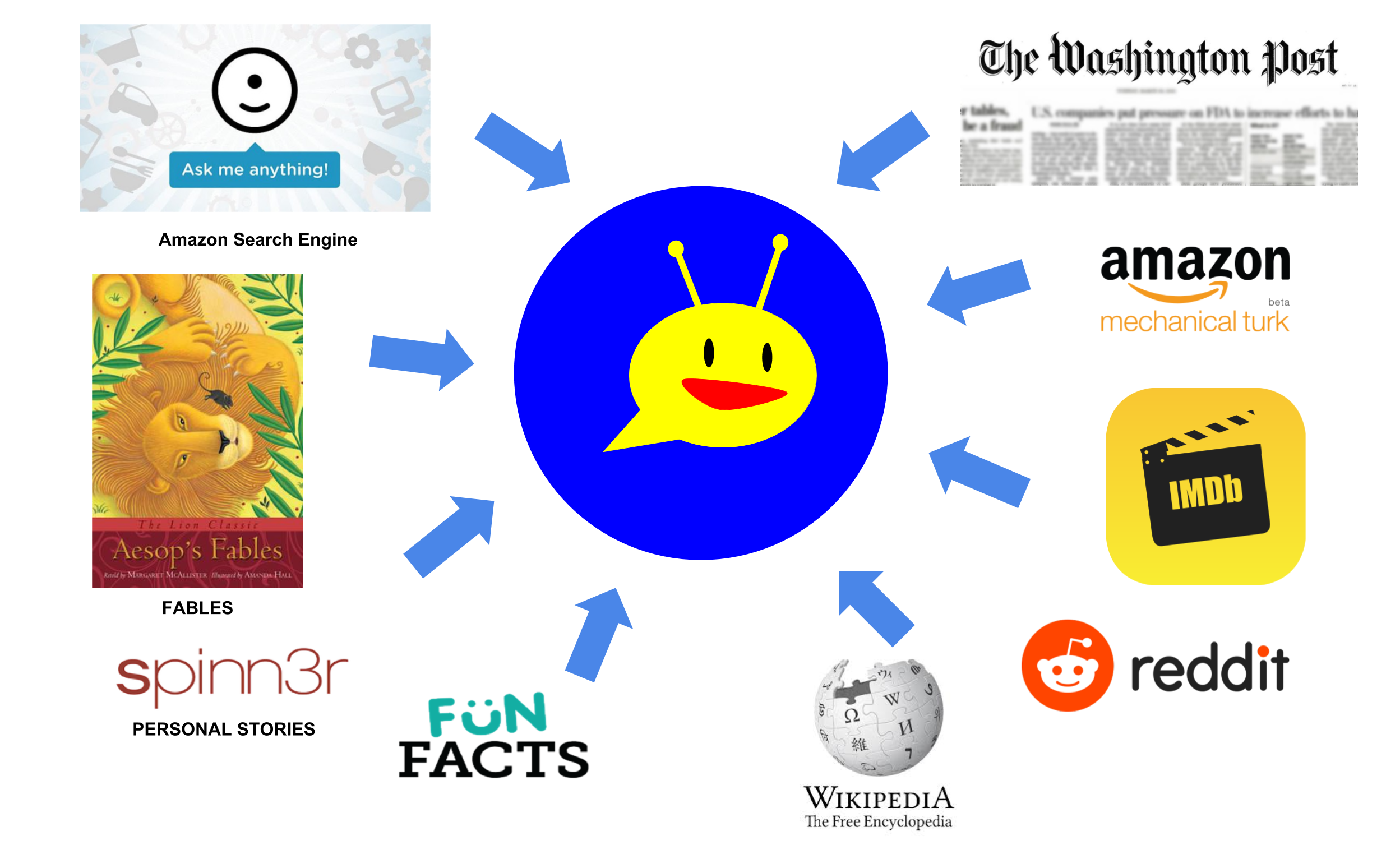}
  \centering
  \caption{Content sources for SlugBot.}
  \label{fig:content}
\end{figure}

One of the greatest challenges of developing an Alexa Prize Bot is the
need to collect and curate content from a wide variety of sources to
cover a wide range of topics of interest. Figure~\ref{fig:content}
specifies how we have sourced content from our own Mechanical Turk
HITs as well as the data collected by the 2017 Alexa Prize Edina team~\cite{krause2017edina}, as well as
sourcing content from specific Reddit subreddits, story corpora such
as Aesop's Fables, trivia websites, and news sources among others. All
of this content needs to be processed for named entities and dialogue
acts in order to be used by our dialogue manager. We also use sections of the Amazon provided FUD and CAPC datasets to motivate our topical and functional expansion. This content is used
in combination with search engines as described below in more detail.

\textbf{UniSlug:} In addition to these content sources, we host the UniSlug graph
database illustrated in Figure~\ref{fig:unislug} on EC2. This is a key
part of our architecture and it must be accessible in real time since
it is used both for NLU and for dialogue management. UniSlug consists
of 275GB of ontological knowledge base data from various sources, consisting of one billion relations and over 100 million
entities. This allows us to flexibly reference and navigate semantic
relations as well as expanding the knowledge that SlugBot uses in the
system initiative modules described in Section \ref{sec:active}.

We distinguish between two types of memory when we actually
utilize this content: short term memory (STM) and long term memory
(LTM).

\textbf{The STM cluster} is responsible for managing our system state,
representing the current discourse context, and handling data which is
localized to specific functionality. This local memory helps us to
improve the efficiency of data access and reduce the workload of
remote databases.  

\textbf{The LTM cluster} is responsible for managing our
corpora and other large datasets. We utilize the Amazon Relational
Database Service to store real-time search data in addition to other curated content that is used by various modules.  

LTM and STM communicate with each other to exchange data: the data which is
out-of-date should transfer into the LTM while the conversationally
relevant data should be made available in the STM. Both memory
clusters are comprised of a network of memory nodes which each have
their own responsibilities.

\subsection{Dialogue Manager}
\label{sec:agent_core}

\begin{table*}[ht!]
\begin{small}
\begin{center}
\begin{tabular}
{p{2cm}|p{1cm}|p{10cm}}
\toprule
\bf Module & \bf Section & \bf Description \\
\midrule
Base Responses & \ref{sec:agent_core} & State-specific responses like handling repeat requests or prompting with a menu.\\ \midrule
Opinions & \ref{sec:reactive} & Solicit, provide, and justify opinions about detected entities. \\ \midrule
Question Answering & \ref{sec:reactive} & Question answering modules including ELIZA, Evi, Wikipedia, and DuckDuckGo. \\ \midrule
Well-being  & \ref{sec:reactive} & Detect user well-being class and take different action accordingly. \\ \midrule
Intimacy  & \ref{sec:reactive} & Gain intimate connection with the user and understand their interests. \\ \midrule
Out-of-Domain & \ref{sec:reactive} & Out-of-Domain responses to sustain the conversation if there are no other good options. \\ \midrule
Retrieval & \ref{sec:retrieval-sec} & Elasticsearch index used to retrieve appropriate responses. \\ \midrule
Storytelling & \ref{sec:active} & Tell the user a story and answer questions based on a corpus of personal narratives. \\ \midrule
Recommendation & \ref{sec:active} & Able to recommend different elements of UniSlug, specifically in the travel domain. \\ \midrule
Recursive & \ref{sec:active} & Facts or trivia triggered recursively until user wants to change topics. Additionally, short 2-turn sequences including riddles and would you rather questions. \\ \midrule
Flow Manager & \ref{sec:flowman} & Managing dialogue flows about a variety of topics (currently, 42) as a way to rapidly increase popular topic coverage. \\ 
\bottomrule         
 \end{tabular}
 \caption{Summarized list of different modules.}
\label{table:modules}
\end{center}
\end{small}
\end{table*}

The dialogue manager is tasked with handling the most basic functionality, such as detecting repeat requests, stop requests, and prompting the user with a menu of topics to help transition into domains with a breadth of coverage. We maximize the number of unique experiences within a single conversation by prioritizing unexplored topics. We found that having a menu made an appreciable difference in conversational quality. Additionally, the dialogue manager will periodically ask the user for feedback on a recent set of turns or apologize for delays in response time. When designing our feedback responses, we made sure to phrase them such that they emphasize improvement for the user, not SlugBot's experience. A potential exchange can be seen in example \ref{table:feedback}. 

\begin{wraptable}{L}{0.58\textwidth}
\begin{small}
\begin{tabular}
{p{0.9cm}|p{6.0cm}}
\toprule
Agent & {So anyways, We just talked about some would you rather questions. If you don't mind me asking, did you have fun talking about it? Would you like to do it again some time in the future?}\\
User & {no} \\
Agent & {I see. I'm bummed you didn't have fun. I'll remember this for the future so that we have more fun next time.}\\
\bottomrule               
 \end{tabular}
\caption{Sample of an interaction using our feedback exchange.}\label{table:feedback}
\end{small}
\end{wraptable}

The dialogue manager also sets system expectations. Specifically, we indicate the data representing the preconditions which must be satisfied in order for a particular action to be valid. Our expectations can be satisfied by observing attributes from the system's state. We can define our expectations using a variety of state variables including direct keyword matching, specific utterance attributes such as dialogue act or sentiment score, or even be contingent on the result returned by a function. 

\subsection{Dialogue Modules}
\label{supplementary_cores}

Table~\ref{table:modules} lists the dialogue modules implemented in our system, with descriptions of each module in the following sections. Our dialogue manager maintains control over the dialogue modules, allowing us to easily change the type of responses which we pool. We feature primarily two module classes, mixed initiative in the case where our agent does not have control of the conversation, and system initiative when the agent is driving the conversation. The ranking algorithm which is used to pick the best candidate module is described in Section \ref{sec:scoring}.

\subsubsection{Mixed Initiative Modules}
\label{sec:reactive}

\begin{wraptable}{L}{.5\textwidth}
\begin{small}
\centering
\begin{tabular}
{p{0.9cm}|p{5.1cm}}
\toprule
User & {what do you think of watchmen} \\
Agent & {I really like watchmen. In fact I think watchmen is action packed because there are lot's of cool fight scenes. How about you, what's your opinion?} \\
User & {do you like watchmen} \\
Agent & {Bot: Yes, I believe watchmen is really interesting. And you? What's your opinion?}\\
User & {What's your favorite color?} \\
Agent & {I really like infrared. It makes me feel warm all over.}\\
User & {I like blue} \\
Agent & {Why do you feel positively about blue?}\\
\bottomrule         
 \end{tabular}
 \caption{Sample of an interaction using our opinion data.}
 \label{table:opinions_ex}
\end{small}
\centering
\end{wraptable}

Mixed initiative modules are designed to solicit an even exchange in content per turn from both the user and the agent. More specifically, this is the case in which the agent is not strictly in control of the conversation and in fact, our content is likely to be a reaction to the user's initiative. Such user initiative could be asking a question, soliciting an opinion, or having general chit-chat outside of a controlled dialogue flow. 

\textbf{Opinions:} Here we can learn more about the user by soliciting their opinion of a contextually relevant entity. Naturally, if we can solicit user opinions, it's important that we are able to provide and justify our own opinions. To accomplish this, we load our agent's profile with opinions about various entities and abstract concepts. The first time a user engages with the system we randomly select opinions to include in the agent's profile, allowing us to take the preliminary steps towards giving the agent a unique and identifiable personality. Our opinion dataset is extracted from online reviews of movies, video games, and books. We have also handcrafted data points which would give us good general coverage of conceptual opinions based on popular topics which would be difficult to extract from any data source, such as ``What is your favorite color?''. Finally, we have also collected full opinions, and justifications of existing opinions using Mechanical Turk. When collecting new opinions we focused on positive, non-controversial topics. In Table \ref{table:opinions_ex} we have synthesized an example which leverages our structured opinion data to answer solicitations from the user.

\textbf{Question Answering:} Our question answering mechanism is a three step inspection of the query. First, if there are not enough content words to detect the intent of the question, we probe for more information using a modified version of ELIZA~\cite{Weizenbaum1966ELIZA}; we found that users generally reacted poorly to some of the more intimate prompts when used out of context. If the system has the initiative, such as when we are telling a story, we assume the associated module will be able to answer questions using the module's structured data. If neither of these conditions are true or we don't yet have an answer, we perform coreference disambiguation and query our search engine. Our search engine consists of three search components which are queried in the following order: Evi\footnote{\url{https://www.evi.com/}}, Wikipedia\footnote{\url{https://www.wikipedia.org/}}, and finally DuckDuckGo\footnote{\url{https://duckduckgo.com/}}. If we have still not returned an answer, we inform the user that at this time we are not able to answer the question. We noticed through anonymous user feedback that, in general, users were okay with our system not being able to answer questions as long as the agent made it clear that it was not sure of an answer. An example demonstrating this module is shown in Table~\ref{table:qa_ex}.

\begin{wraptable}{L}{0.5\textwidth}
\begin{small}
\begin{tabular}
{p{0.9cm}|p{5.1cm}}
\toprule
User & {What is the capitol city of Mexico.}\\
Agent & {The capitol city of Mexico is Mexico City. (search)} \\
User & {What is it's population?}\\
Agent & {The population of Mexico City is 8.8 million. (disambiguation + search)} \\
User & {Okay, how is it that you are smart?}\\
Agent & {Why do you think I am smart? (ELIZA)}\\
User & {Just a guess. Tell me a story.}\\
Agent & {Did I ever tell you one time my pet Moses really scared me? (enter storytelling)}\\
User & {No, what kind of pet is it?}\\
Agent & {Moses is a chinchilla. (storytelling answer)}\\
\bottomrule               
 \end{tabular}
\end{small}
\caption{Sample of an Interaction Using our Question Answering}\label{table:qa_ex}
\end{wraptable}

\begin{figure}
\centering
\begin{floatrow}
\capbtabbox{%
  \begin{tabular}
{@{}p{1.5cm}|p{3.5cm}@{}}
\toprule
Type & User Utterance  \\ \midrule
Positive & {I'm doing good.}  \\
Negative & {Not good.}  \\
Bored & {Bored.}  \\
Offended & {Say sorry!}  \\
Hostile & {You \#\&\%!}  \\
\bottomrule         
 \end{tabular}
}{%
  \caption{Sample user responses when agent asks "how are you doing?"\label{table:user_mood}}%
}
\capbtabbox{%
  \includegraphics[scale=.22]{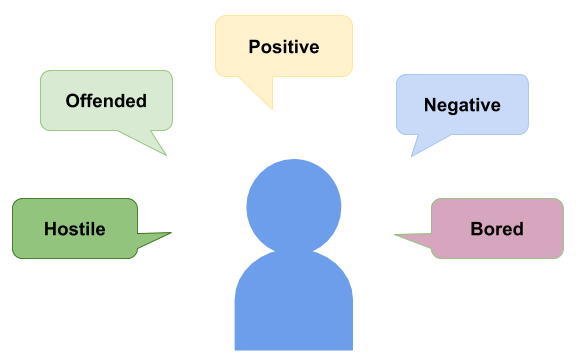}
}
{%
  \captionof{figure}{Major user moods.\label{fig:user_mood}}%
}
\end{floatrow}
\end{figure}

\textbf{Well-being:} Our well-being module is designed to understand the user's basic mood through sentiment analysis and a series of questions. We learn from the conversation history that, besides positive and negative, the major emotions which effect our dialogue strategy would also include: neutral, bored, hostile, and offended. Table~\ref{table:user_mood} and Figure~\ref{fig:user_mood} show the sample user utterances after our agent asks "How are you doing?". The \emph{Bored} users tend to say "I don't know" when the agent asks other questions, the \emph{Offended} users are very sensitive to personal questions, the \emph{Hostile} users are aggressive and tend to give us low rating. Therefore, our agent would take different actions accordingly. For example, we avoid asking to many question to the \emph{Bored} users but give more direct suggestions. We avoid asking further personal questions to the \emph{Offended} and \emph{Hostile} users. As for the \emph{Positive} users, we would let them lead the conversation, while for the \emph{Negative} users, we would suggests some resources in our dialogue system to improve their well-being.

\textbf{Intimacy:}
In addition to learning more about the user's well-being, SlugBot attempts to gain a more intimate connection with the user by asking intimacy questions and understanding their interests. Our intimacy questions are handled using a similar strategy as our other 2-turn recursive questions. Each question has been annotated with an intimacy level, which we use as a filter when curating content. As our conversation with the user progresses, we relax this constraint allowing SlugBot to ask more intimate questions. We also try and gauge the user's interests based on our unexplored system modules, and the user's context. The former case allows us to surface new content, while the latter allows us to leverage the context of the conversation to gain further understanding of the user and surface contextually relevant turns.

\begin{wraptable}{L}{0.5\textwidth}
\begin{small}
\begin{tabular}
{p{0.9cm}|p{5.1cm}}
\toprule
Agent & {What characteristics of yourself are you most proud of?}\\
User & {My face.} \\
Agent & {I see, for me personally you might ask? I have quite speedy natural language processing capabilities, coupled with a vast range of different conversational abilities and actions that I can perform for you. Would you be okay answering another question?}\\
User & {No} \\
Agent & {Okay, are you interested at all in dinosaurs?}\\
\bottomrule               
 \end{tabular}
\end{small}
\caption{Sample of interaction using our intimacy modules.}\label{table:intimacy_ex}
\end{wraptable}

\subsubsection{Dialogue Management Driven Retrieval} 
\label{sec:retrieval-sec}

Our retrieval module is constructed using an
Elasticsearch\footnote{\url{https://github.com/elastic/elasticsearch}}
based index, which has advantages of fast full text search capability
and scalability. We built our indices on dialogue based, fact based
and social media data with different search criteria. 

An additional data sources, tailored towards a two-party open domain
conversation, is a crowd-sourced corpus of conversational turns
on a variety of topics using qualified workers on Mechanical
Turk. From this data collection, we gathered 17,364 topic annotated
turns. In addition to our crowd-sourced data collection we further
supplemented our index by handcrafting responses which would have
only been captured by observing them in our user trials, as discussed
further in Section \ref{sec:discussion}. 

We also incorporate trivia, fun
facts, jokes, would you rather questions and hypothetic questions,
consisting of 5,439 high quality utterances, with Elasticsearch index,
it retrieves utterances when the users are interested in specific
topics or entities. 

We also utilized the Reddit
API\footnote{\url{https://github.com/praw-dev/praw}} selectively
collecting from 17 subrreddits resulting 38,172 Reddit posts. We built
indices and customized queries based on the topic, keywords, entities
and meta fields. Moreover for better quality of responses using Reddit
posts, we examined the data and applied filters to filter out posts
that are not appropriate for dialogue turns. We also used the corpus
that Ben Krause et. al. published as part of their Edina project~\cite{krause2017edina}. We extracted every utterance of their published data and applied our own set of filters.

We first ran the utterances through our own pipeline that determined
the topic based in part on Google knowledge graphs. Our queries
then used our filters to select responses based on topic. We removed
utterances that contained any type of anaphora we could identify. This
included pronouns, specific temporal indicators, and agreement indicators. 
Sentences with pronouns could be referring to someone or something that occurred in a prior utterance. Specific times are also potentially problematic because events that occurred at a specific time, such as a game or a concert,
do not generally reoccur at the same time. We also carefully examined
the utterances to develop subtle rules for removing potentially
confusing or inappropriate comments.

\begin{wraptable}{L}{0.5\textwidth}
\begin{small}
\begin{tabular}
{p{0.9cm}|p{5.1cm}}
\toprule
User & {What is your favorite Beatles album?}\\
Agent & {My favorite album is "Sargent Pepper's Lonely Hearts Club Band", however "Abby Road" is a close second. (Edina)} \\
User & {I liked Abby Road better than Sargent Pepper.}\\
Agent & {I feel like Abbey Road was every bit as chaotic as White Album, yet excellent audio engineering was what truly made it feel cohesive. (Reddit)}\\
User & {How many copies did Sargent Pepper sell?}\\
Agent & {As of 2011, it has sold more than 32 million copies worldwide, making it one of the best-selling albums.  (disambiguation + search)} \\
\bottomrule               
 \end{tabular}
\end{small}
\caption{Sample of an interaction using our Edina and Reddit data.}\label{table:red_ed_ex}
\end{wraptable}

\textbf{Out-of-Domain:} Using a combination of these modules we also designed our out-of-domain responses for when we are not confident in any specific answer. For example, if we are able to detect an entity but do not have a direct follow-up utterance, we can either give an opinion, ask the user for more information, verify the entity by asking about one of its related synonyms, or use our search engine to try and retrieve some possible summary-based response. If our system does not have above 0.8 confidence in our top-ranked answer, we employ a transition strategy which attempts to engage the user with an unexplored module or topic. A sample fallback utterance assumes we have not yet discussed the user's well-being and is as follows: \textit{I'm really sorry about this, but I'm not sure what to say next. My robot brain is telling me I'm not quite sure how to respond to what you just said. Do you mind if I switch to a new topic? How is your day going?}

\subsubsection{System Initiative Modules}
\label{sec:active}

Our system initiative modules gain control of the conversation by assuming the role of content provided and driving the conversation. These modules are designed to encourage extended multi-turn interactions by routing the user back to the module when the conversation starts to stray. 

\begin{wraptable}{L}{0.5\textwidth}
\centering
\begin{small}
\begin{tabular}
{p{0.9cm}|p{5.1cm}}
\toprule
Agent & {Do you want to hear some science facts?} \\
User & {Yes} \\
Agent & {Did you know that At over 2000 kilometers long, The Great Barrier Reef is the largest living structure on Earth. Want to hear another?} \\
User & {Sure why not} \\
Agent & {How about this one. The average human body carries ten times more bacterial cells than human cells. Want to hear more?} \\
\bottomrule               
 \end{tabular}
\end{small}
  \caption{Sample of an interaction using recursion on the Science topic.\\}\label{table:recursive_ex}
\end{wraptable}

\textbf{Stories:} Stories are often told conversationally, and in a two-party setting it is natural that one speaker will assume the responsibility of content provider while the other will assume the role of recipient. This can lead to an entertaining experience for the recipient, while also minimizing the conversation's reliance on their content. Therefore, if we are able to replicate this environment, we hope to create an engaging multi-turn experience. As described in previous iterations of SlugBot~\cite{bowden2018slugbot}, we incorporated a large corpus of personal blogs~\cite{Personabank}, from which we built a small dataset of around 40 narratives which are interesting and do not contain inappropriate content. After closely inspecting the results of our preliminary evaluation, we realize that this content was not received highly as anticipated. We attribute this to 1) the content being boring, or so unrealistic for an Echo device's personal narrative, and 2) improper control of the stopping points throughout the story led to random pauses and allowed the user to ask more questions which we could not answer.

To address these issues we feature three different sources of story content this year. Fables, dreams that SlugBot had, and SlugBot's adventures as an Echo device. Our fables content is provided by a large corpus of annotated versions of Aesop fables~\cite{elson2012dramabank}. Both the dreams and the personal adventures data we crowd-sourced using Mechanical Turk. All three of these data sources allow the user to sufficiently suspend their disbelief that such an embodiment could be in the story, and since we targeted popular entertainment franchises and topics when curating the data, we believe it is much more engaging than our previous implementation. Further, we have annotated our stories in chunks, where each chunk ends with a natural pausing point. We appended tag questions to the end of these chunks in order to implicitly discourage the user from asking us hard questions we are not capable of answering. Finally, we used SSML markup to fix issues with timing and prosody in addition to giving the stories a sense of dramatic flare.

\textbf{Recommendation:} In daily life, conversation on topics such as movies, books, travel etc., people would like to recommend and discuss similar movies, another related book, or other interesting places that they visited while traveling. This inspires us to give recommendation on such topics so that it not only improves the depth of the conversation but also makes a better user experience. To be able to understand how entities are related and retrieve such information, we use UniSlug, as mentioned in Section~\ref{sec:data}. We first identify the entities mentioned, then we explore adjacent vertices in the graph by various relations. 

Our primary implementation of this feature is in the travel domain. For example, if the user wants to travel to New York, we can recommend some famous places in New York. Or if the user plans to visit the Eiffel Tower, we can recommend some other tourist spots that are also located in the same area. Based on the relations in the graph database, we are able to curate our responses most appropriately. 

\begin{table}
\begin{small}
\begin{tabular}
{p{0.9cm}|p{11.1cm}}
\toprule
Agent & {How about I ask you some would you rather questions?}\\
User & {Okay} \\
Agent & {Would you rather read books that are nonfiction or would you rather read a steamy romantic novel?}\\
User & {nonfiction} \\
Agent & {For me personally? I would rather curl up underneath a blanket with a physical copy of a book while sitting on the couch, with a warm cup of tea. It just makes me feel so relaxed and I can forget about my troubles for a while. Do you want to hear another books question?}\\
User & {No talk about video games}\\
Agent & {If you could be an character from any video game who would you be and why would you pick them?}\\
User & {mario from super mario odyssey cause you can throw his hat and go into to that character that you thought you had on}\\
Agent & {Cool, for me personally, Nathan drake because he gets to go on lots of cool adventures and has lots of awesome weapons Do you want me to ask another video games question?}\\
\bottomrule               
 \end{tabular}
\end{small}
\caption{Sample of an Interaction Using our 'Would You Rather' Sequence.}\label{table:sequence_ex}
\end{table}

\textbf{Recursive:} Finally, we have a set of modules which acts recursively. Here we can inform the user of various headlines from new sources, or give the user facts about a topic of their choice. We are able to recurse over this functionality by simply continuously giving them information until they explicitly transition out of the recursion. We have included an example of this in Table \ref{table:recursive_ex}. We can also create recursive 2-turn sequences by asking the user a sequence of riddles, would you rather questions, and hypothetical questions. All three of these cases allow the user and agent to converse for a couple of turns briefly about the sequence before recursing, an example of this can be seen in Table \ref{table:sequence_ex}. We collected approximately 1,500 question/answer pairs using crowd-sourced labor across each of the topics supported by the flow manager, as discussed in Section \ref{sec:flowman}. These recursive loops are highly effective at keeping the user engaged with the agent in a multi-turn context without having to worry about a complex dialogue flow. Moreover, having topic-annotated content allows us to fluidly connect related content from SlugBot's modules. 

\subsection{Flow Manager}

One of the primary modules in our system, the flow manager, is responsible for managing the flow of dialogue related to a given topic or utility. A flow, as seen in Figure 2, is organized in a graph structure where each node has specific preconditions, postconditions, and actions which work together to sustain a natural dialogue about any particular subject. While flows in general are meant to provide coverage of some specified root topic, like Books, it's important to note that many flows contain multiple subroots, such as "favorite genre", "book trivia", or "bestsellers". The user can directly trigger a flow about a given topic by using related keywords, or by expressing interest in the flow if the prompt is selected by the system when propositioning topics.

\label{sec:flowman}
\begin{figure}
\centering
\begin{floatrow}
\capbtabbox{%
  \includegraphics[scale=.4]{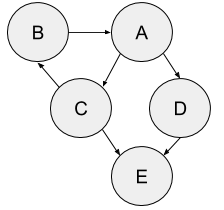}
  \label{fig:sample_flow}
}
{%
  \captionof{figure}{Sample flow.}%
}
\capbtabbox{%
  \begin{tabular}
{@{}p{0.1cm}|p{0.9cm}|p{4cm}@{}}
\toprule
1 & User & {Precondition A} \\
2 & Agent & {Action A, Postcondition A} \\
3 & User & {Precondition C} \\
4 & Agent & {Action C, Postcondition C} \\
5 & User & {Precondition B} \\
6 & Agent & {Action B, Postcondition B} \\
7 & User & {No Precondition} \\
8 & Agent & {Exit Flow} \\
\bottomrule         
 \end{tabular}
}{%
  \caption{A sample conversation.}%
}
\capbtabbox{
\begin{tabular}
{@{}p{0.1cm}|p{2cm}@{}}
\toprule
1 & {A, .... ,Z} \\
 & \\
3 & {C, D} \\
 & \\
5 & {B, E} \\
 & \\
7 & {A} \\
 &  \\
\bottomrule         
 \end{tabular}
}{%
  \caption{Expecting.}%
}
\end{floatrow}
\end{figure}

\begin{wraptable}[20]{L}{0.71\textwidth}
\begin{small}
\begin{tabular}
{c c c}
\toprule
&Flow Topics&\\
\hline
Music&Fun Facts&Harry Potter\\
Technology&Trivia&Star Wars\\
News Headlines&Hobbies&Star Trek\\
Box Office&Holidays&Monsters\\
Sports&Astronomy&Marvel Cinematic Universe\\
Fashion&Animals&Pokemon\\
Shopping&History&Cartoons\\
Travel&Board Games&Fictional Characters\\
Nutrition&Books&Tolkien\\
Health&Language&Science Fiction\\
Favorite Food&Famous Quotes&Comic Books\\
Recipe&Poems&Pirates\\
Gossip&Weather&Video Games\\
TV&Horoscope&Dinosaurs\\
\bottomrule 
 \end{tabular}
 \centering
 \caption{List of the Current Flows Supported by SlugBot.}
 \label{table:flows}
\end{small}
\end{wraptable}

Our preconditions are represented by the expected data discussed in Section~\ref{sec:agent_core}. Postconditions can represent a variety of desirable effects which only occur after the response has been realized by the dialogue manager. It is within these postconditions that we can indicate calls to external functions or update specific state variables. Finally, there are actions which also occur. These actions can modify a candidate utterance or delegate responsibility for response curation to a different module.

Flows represent a high-level abstraction of our entire system's functionality, allowing a new designer to rapidly add content to the system without needing to familiarize themselves with the underlying architecture. As seen in Table~\ref{table:flows}, our system currently supports 42 flows covering a diverse range of topics. Since these flows represent a high-level abstraction of the entire system, we found that reusing successful modules is an effective means of bootstrapping flows with minimal effort. Specifically, most flows have some recursive trivia-based prompt in their list of subroots. We also found that generically discussing user preferences and utilizing search methods increased the breadth of a flow, while a combination of all methods could increase the depth of a flow.

Another means of increasing the conversation depth, used for example in the Nutrition flow, is by preparing a hierarchical knowledge base that is recursively navigated by the flow. The Nutrition flow uses such a knowledge base of arguments supporting various nutrition facts extracted from Healthline Nutrition\footnote{https://www.healthline.com/nutrition}, with many references across different nutrition topics. The flow starts by offering a controversial fact to the user, and then, based on the user's reaction (such as agreeable, inquisitive, or negative), it responds with a supporting argument, counter-argument, another related nutrition fact, or possibly an information related to the user's follow-up question. In the spirit of the DRDM's method generalizability, this recursive flow design, paired with a suitable knowledge base, can be easily reused for other topics.

\subsection{Ranking}
\label{sec:scoring}

Once we have established a pool of responses, we re-rank them to find a response we are most confident in. Each response has a confidence score which ranges from 0 to 1. Each response is assigned a base confidence value which is increased and decreased based on contextually information. We have established a base confidence value of 0.6, which is a value we have carefully selected to work best in our system. While some of the following metrics are automatically calculated, this base confidence in addition to the weights assigned to each metric are manually assigned.

If the user says \textit{I like video games} for example, the \textit{video game} conversation starter will have a confidence of 1, because our system is very confident it is the best next response. If the user said \textit{I like dogs}, the \textit{video game} prompt would have a confidence of 0.6, indicating it as a valid topic starter, but only if we have nothing more relevant to say. At this stage, our sensitive content filter will invalidate any response with explicit content and detect if a priority response has been triggered. Priority responses are valid regardless of our current state and indicate responses which are to be uncontested - such as repeat requests or stop session markers. Finally, for all other responses we update their confidence using Equation \ref{eq:scoring}. We attempt to increase our confidence in the response by looking for contextually relevant content and inspecting the current system state. Our context score is calculated by considering overlapping content words and entities in addition to our system's state variables.  

Equation \ref{eq:loss} represents how we penalize a given response. A response is in a state of incoherence if it does not belong to the current system initiative module. For example, if we are playing a specific game, and a response stems from anywhere besides that game it would create incoherence within the conversation. In order to maintain module coherence we apply an empirically derived 0.15 penalty to these responses. While we leverage the state tracking done by our Short Term Memory to avoid repetitious utterances, some general prompting phrases such as {\it ``would you like to play a game?''} are still valid despite being already said. In order to increase the diversity of a user's experience without limiting the variety in our response pool we apply a 0.05 penalty to prompts which have already been explored. Furthermore, we noticed from our own experimentation that long utterances from mixed initiative modules tended to be received poorly. An example of this includes long news headlines and overly verbose indexed responses, where we have limited control over the phrasal timing. We therefore applied a length based penalty to these utterances. 

\begin{equation} 
\label{eq:scoring}
\textbf{def score:  } r_{i}.confidence = min(max(context(r_{i}), r_{i}.confidence) - loss(r_{i}), 1)
\end{equation}

\begin{equation} \label{eq:loss}
\textbf{def loss:  } r_{i}.confidence = incoherence(r_{i}) + repeat(r{i}) + sentLen(r_{i})
\end{equation}

In the case that more than one response shares the highest confidence score, we resolve ties by randomly choosing among the candidates. It should be noted that in most cases candidates which are tied for the maximum score all tend to realize the same intention.

\subsection{Postprocessing}
\label{sec:postprocess}
\textbf{Stylistic Variation:} When realizing our top response candidate, we use a statistical natural language generation engine \cite{bowden2017monolog} to add variation in a way which seems to increase the coherence of the response. 
Our most common use case is prepending our response with an acknowledgement hedge - enabling a more natural transition between topics. For example, when the user is giving an opinion, the agent's response will begin with ``I see'' or ``Oh really''. Additionally, we gathered specific user intents by analyzing our logged data to add specific hedges to the start of the utterance when appropriate. For example, if the user says ``thank you'' or ``are you kidding'' we will prepend our response with ``You're welcome'' and ``I kid you not'' respectively.

\textbf{Response Merging:} It is commonly the case that multiple good responses will be scored highly by our system. With certain types of content it is possible to merge these responses to give SlugBot a richer response. After reading through the user data, it seems clear that always ending our utterance with a question retains some control over the conversation and invokes follow up content from the user. Therefore, we will frequently merge a fact or statement with a question about the same topic or entity. An example of this output can be seen in Table \ref{table:merged_ex}. 

\begin{table}
\centering
\begin{small}
\begin{tabular}
{p{2cm}|p{10cm}}
\toprule
Source & Candidate\\
\hline
Wiki & {The Matrix is a 1999 science fiction action film written and directed by The Wachowskis, starring Keanu Reeves, Laurence Fishburne, CarrieAnne Moss, Hugo Weaving, and Joe Pantoliano.}\\
Trivia & {In the first forty-five minutes of the film, Neo has eighty lines. Forty-four of these lines are questions, just over half of his total dialogue, averaging at roughly one question per minute.}\\
Opinion & {What are your thoughts about the matrix?}\\
Intimacy & {Why are you interested in the matrix?}\\
Index & {Did you enjoy the follow up movies to the matrix trilogy?}\\
\hline
\hline
Agent & {The Matrix is a 1999 science fiction action film written and directed by The Wachowskis, starring Keanu Reeves, Laurence Fishburne, CarrieAnne Moss, Hugo Weaving, and Joe Pantoliano. What are your thoughts about the matrix?} \\
Agent & {The Matrix is a 1999 science fiction action film written and directed by The Wachowskis, starring Keanu Reeves, Laurence Fishburne, CarrieAnne Moss, Hugo Weaving, and Joe Pantoliano. Why are you interested in the matrix?} \\
Agent & {In the first forty-five minutes of the film, Neo has eighty lines. Forty-four of these lines are questions, just over half of his total dialogue, averaging at roughly one question per minute. Did you enjoy the follow up movies to the matrix trilogy?} \\
\bottomrule               
 \end{tabular}
\end{small}
  \caption{Sample of merged utterances for the input "Tell me about the matrix".\\}\label{table:merged_ex}
\end{table}

\textbf{SSML:} At this stage we also apply any SSML markup which is encoded in the response. We found that SSML seemed to improve certain sections of dialogue extremely well. Specifically, we primarily used it for 1) adding additional pauses in sentences, 2) correcting the way certain words are interpreted by the TTS engine, 3) adding dramatic flare to the various stories our system is capable of telling.

\subsection{Web Application Architecture}
\label{sec:amazon_infrastructure}

\begin{figure}[ht]
  \includegraphics[scale=.15]{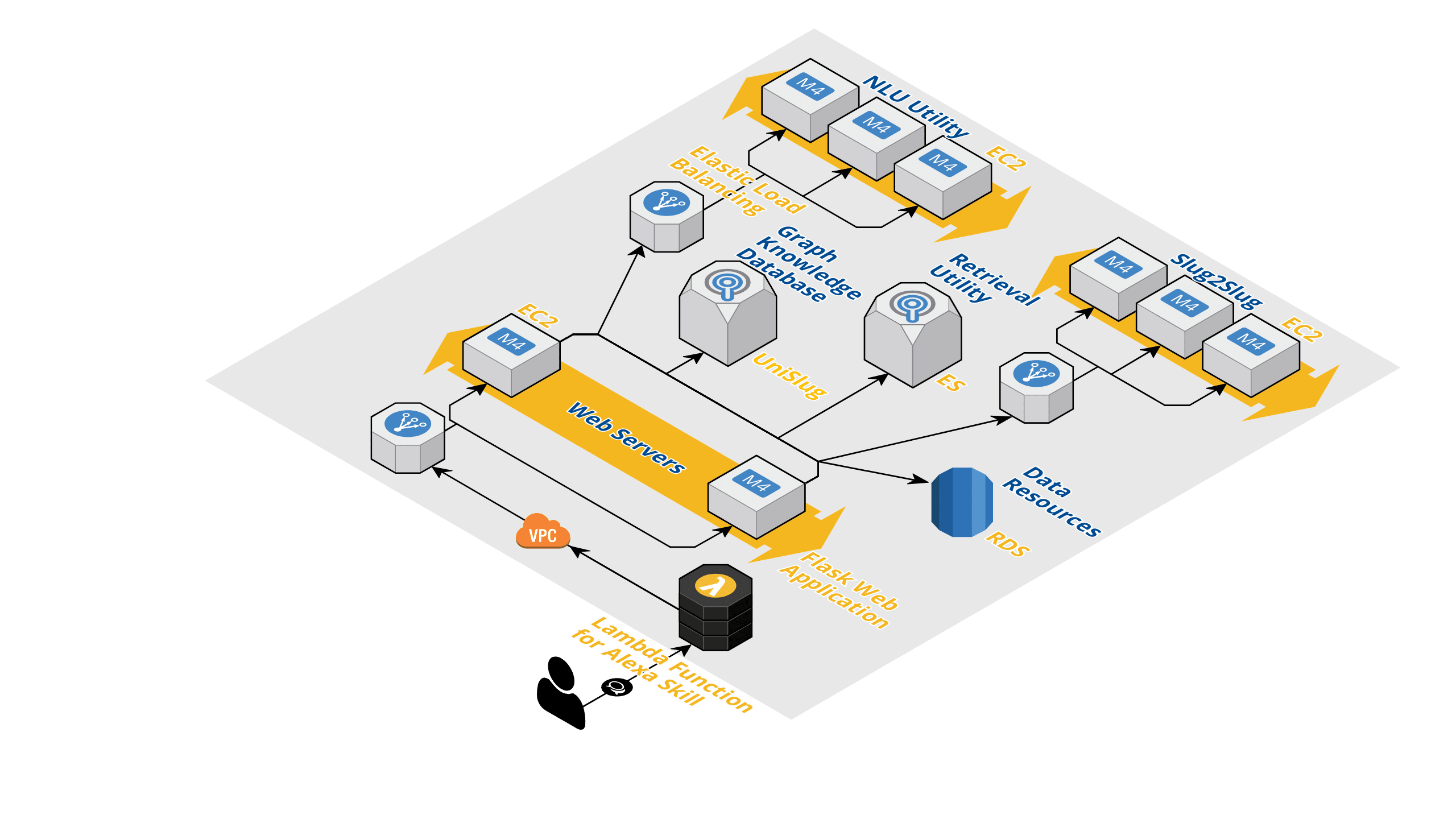}
  \centering
  \caption{Our Web Application Architecture.}
  \label{fig:webapp_arch}
\end{figure}

Figure ~\ref{fig:webapp_arch} depicts the architecture of our web server in details.

\textbf{Flask Web Application:} We apply the Flask web application structure to facilitate communication between our framework and AWS Lambda. We consider exception handling an important part of our system design. Not only do we keep the daily user interaction logs in EC2 for manual inspection, but we also found the CloudWatch Alarms useful for error notifications.

\textbf{Elastic Load Balancer:} Elastic Load Balancer was easier for us to incorporate in our existing framework than Elastic Beanstalk, even though Elastic Beanstalk is convenient for deployment. To validate our load balancing, we use the Locust\footnote{http://locust.io/} load testing framework to expose potential hazards.

\textbf{Natural Language Understanding (NLU):} Our NLU utilities, specifically CoreNLP, runs on its own server and could represent a performance bottleneck especially in the case of long utterances. Therefore we run these NLU utilities on a separate EC2 instance which communicates with the web server.

\textbf{Data Resources:} We use a vanilla RDS Relational Database (Aurora MySQL) with only one replication to store our data resources. We create index for the frequent query table to reduce latency. 

\textbf{UniSlug Graph Knowledge Database}: Our primary knowledge base, UniSlug, is running on a separate EC2 instance to increase the performance of real time queries.

\textbf{Retrieval Utility}: We are running an Elasticsearch server on a separate EC2 instance to support our retrieval based modules and future reinforcement learning in real time. 

\textbf{Natural Language Generation (NLG)}: We have a set of EC2 instances which are serving our neural language generation model, Slug2Slug. We use multiple instances in order to query and ensemble the results of multiple models as quickly as efficiently as possible.

\section{Discussion}
\label{sec:discussion}

\begin{table}
\begin{small}
\centering
\begin{tabular}
{c|c|c|c|c}
\toprule
Module Class&Average User Rating& \# User Turns& \# Unique Conversations&Average Length\\
\hline
Mixed-Initiative&3.16&38856&21346&1.37\\
Retrieval&3.08&11722&7353&1.32\\
System-Initiative&3.58&49571&7853&8.43\\
Flow Manager&3.35&52472&6274&8.36\\

\bottomrule         
 \end{tabular}
 \caption{An evaluation of the four different classes of features in SlugBot. Here our numbers are only considering the user turns.}
 \label{table:class_eval}
\end{small}
\centering
\end{table}

In Table \ref{table:class_eval} we present an evaluation of our four different classes of dialogue modules as described in Sections \ref{sec:reactive} - \ref{sec:flowman}. This data was based on the user feedback received from August 1st to August 28th, for which our system's average user rating was 3.31.

As predicted, when we relinquishes more control to the user it becomes harder to respond appropriately. This is reflected by the fact that our average conversational score for both mixed-initiative and retrieval-based models is lower than our average system performance. We also notice that both strategies are only used slightly more than 1.3 user turns per unique conversation. This makes sense, as providing continuous follow-up content to retrieved or search-based content can be difficult.

Our highest rated class of responses originate from our system-initiative modules, as described in Section \ref{sec:active}. Moreover, we see that on average a conversation which utilized these modules contained more than 8 user turns of related content, which represents multiple games or stories told in the same conversation. This indicates that while the user is individually contributing less overall to the dialogue, driving a conversation with topically relevant content in an entertaining fashion is still an effective dialogue strategy capable of engaging users for a multi-turn conversation.

Finally, we look closer at our flow manager, which is responsible for sustaining general chit-chat conversation about the 42 different topics described in Section \ref{sec:flowman}. Our flow manager received an average score of 3.35, and was similarly able to surface this content for over 8 user turns on average. This indicates to us that our flow management scheme is an effective dialogue management policy. Moreover we'll note that creating new flows is a streamlined process, allowing us to easily scale and create baseline conversations about low data topics, such as famous monsters or dinosaurs. 

As noted in the previous iteration of this competition, it is clear that users still desire that the bot assumes both the responsibility of a personal assistant and  a conversational partner~\cite{bowden2018slugbot}. Users want to adjust the volume of the device, play a specific song, or perform other standard Echo skills. If this type of fluid integration were possible, it would yield an improved overall experience. 

Another issue that we experienced was responding to some user intents. When the user is hostile, it can be difficult, if not impossible, to come up with a good response that the user enjoys. Some user intents are currently off-limits. For example, many users seemed to want Alexa to engage in inappropriate conversations. We obviously cannot just switch to an R-rated dialogue because of a keyword, which may not even be the user intent. Many other users have very strong opinions about the current administration and wish to discuss it, but we cannot give any opinions about it in order to avoid offending people with the opposite view. Some users want to push the artificial intelligence boundaries by saying things like, "I love you." or "You are stupid." Responses to these and other non-task oriented utterances are not easily handled through conventional dialogue systems. We had to make specific responses for these intentions.

Sometimes, the confidence of the user's utterance from the ASR is low, and our system does not know how to respond to an utterance if the confidence is below a certain threshold. This threshold changed throughout the competition as we tried to find the right balance of giving a response which might not be appropriate and not asking the user to clarify their statement.

On the other end of the NLP pipeline, we experimented with a deep-learning approach to language generation. Slug2Slug~\cite{juraska2018slug2slug}, our sequence-to-sequence model that we upgraded to use the Transformer~\cite{vaswani2017attention} architecture, is trained to generate natural utterances from structured meaning representations (i.e. lists of attribute-value pairs). This, however, limits the model to more task-oriented sections of the conversation, which comprise a rather small fraction of an average open-domain interaction with a user. Although we were able to adapt the model to multiple domains through data collection and transfer learning, it would have required considerably more data to make the model robust and versatile enough to be used ubiquitously in place of templates. In order for the system to be able to utilize the model outside of task-oriented domains, we considered implementing a universal meaning representation factory for the system's utterances, which, however, would require even more data annotation. The inference was feasible in real time, but ultimately, due to the above limitations in usability, we decided to forgo the deep learning model and continue using templates. The latter currently provides more personalization and customization power, as well as more flexibility when adding the support for new domains to the system.

\section{Future Work and Conclusion}
\label{sec:future_Work}
In this paper, we have presented our expansion of a scalable socialbot framework and our contribution to the 2018 Amazon Alexa Prize Competition. Additionally, we have described a new dialogue model, DRDM, which is uniquely defined for this task. We have further expanded on our flow-based representation of dialogue, with the intention of increasing its scalability in future iterations of SlugBot.

To further increase the performance of our re-ranker, we incorporate reinforcement learning based on the user feedback in the Alexa Prize data. We defined our decision-making problem as a Markov Decision Process  problem. We trained our reinforcement learning model on different subsets of states so as to find the most efficient set of states for real-time ranking. Currently, we are evaluating the results of this process to determine the optimal policy. 

We have adapted SlugBot's dialogue strategy to provide more user-centric functionality, specifically designed to gauge the interest and well-being of the user. We feel that establishing a more intimate relationship between our socialbot and the user are the next steps towards creating a better, more human-like conversation. User's emotions, such as being happy, sad, or bored, can have a strong influence on the user's interaction with any conversational agent. Adapting and responding to these emotions will more humanize the conversational agent and increase user engagement and satisfaction. Hostile or angry emotions need a different set of responses, which we are still exploring. We are currently working to increase our understanding and ability to adapt and respond to human emotions with the goal of eventually being able to empathize and sypathize with the user as well as provide insightful and meaningful comments to improve the emotional state of the user.

\bibliographystyle{plain}
\bibliography{lyn_ap_2017} 

\begin{thebibliography}{10}

\bibitem{AP80}
James~F. Allen and C.~Raymond Perrault.
\newblock Analyzing intention in utterances.
\newblock {\em Artificial Intelligence}, 15:143--178, 1980.

\bibitem{Ameixa2014outdomain}
David Ameixa, Luisa Coheur, Pedro Fialho, and Paulo Quaresma.
\newblock Luke, i am your father: Dealing with out-of-domain requests by using
  movies subtitles.
\newblock In Timothy Bickmore, Stacy Marsella, and Candace Sidner, editors,
  {\em Intelligent Virtual Agents}, pages 13--21, Cham, 2014. Springer
  International Publishing.

\bibitem{asher-lascarides-book03}
Nicholas Asher and Alex Lascarides.
\newblock {\em Logics of conversation}.
\newblock Cambridge University Press, 2003.

\bibitem{Banchs2012iris}
Rafael~E. Banchs and Haizhou Li.
\newblock Iris: A chat-oriented dialogue system based on the vector space
  model.
\newblock In {\em Proceedings of the ACL 2012 System Demonstrations}, ACL '12,
  pages 37--42, Stroudsburg, PA, USA, 2012. Association for Computational
  Linguistics.

\bibitem{bellegarda2013large}
Jerome~R Bellegarda.
\newblock Large-scale personal assistant technology deployment: the siri
  experience.
\newblock In {\em INTERSPEECH}, pages 2029--2033, 2013.

\bibitem{bowden2017monolog}
Kevin~K. Bowden, Grace~I. Lin, Lena~I. Reed, and Marilyn~A. Walker.
\newblock {M2D:} monolog to dialog generation for conversational story telling.
\newblock {\em CoRR}, abs/1708.07476, 2017.

\bibitem{bowden2018opendomain}
Kevin~K. Bowden, Shereen Oraby, JiaQi Wu, Amita Misra, and Marilyn~A. Walker.
\newblock Combining search with structured data to create a more engaging user
  experience in open domain dialogue.
\newblock {\em CoRR}, abs/1709.05411, 2017.

\bibitem{bowden2018slugbot}
Kevin~K Bowden, Jiaqi Wu, Shereen Oraby, Amita Misra, and Marilyn Walker.
\newblock Slugbot: An application of a novel and scalable open domain socialbot
  framework.
\newblock {\em arXiv preprint arXiv:1801.01531}, 2018.

\bibitem{bowden2018slugNERDS}
Kevin~K. Bowden, JiaQi Wu, Shereen Oraby, Amita Misra, and Marilyn~A. Walker.
\newblock Slugnerds: {A} named entity recognition tool for open domain dialogue
  systems.
\newblock In {\em Proceedings of the Eleventh International Conference on
  Language Resources and Evaluation, {LREC} 2018, Miyazaki, Japan, May 7-12,
  2018.}, 2018.

\bibitem{carlsonMarcuOkurowski01}
Lynn Carlson, Daniel Marcu, and Mary~Ellen Okurowski.
\newblock Building a discourse-tagged corpus in the framework of rhetorical
  structure theory.
\newblock In {\em Proc. of the Second SIGdial Workshop on Discourse and
  Dialog}, Aalborg, Denmark, 2001.

\bibitem{Charrasetal16}
Franck Charras, Guillaume~Dubuisson Duplessis, Vincent Letard, Anne-Laure
  Ligozat, and Sophie Rosset.
\newblock Comparing system-response retrieval models for open-domain and casual
  conversational agent.
\newblock In {\em Second Workshop on Chatbots and Conversational Agent
  Technologies (WOCHAT@ IVA2016)}, 2016.

\bibitem{Duplessisetal16}
Guillaume~Dubuisson Duplessis, Vincent Letard, Anne-Laure Ligozat, and Sophie
  Rosset.
\newblock Purely corpus-based automatic conversation authoring.
\newblock In {\em LREC}, 2016.

\bibitem{DuplessisLLR16}
Guillaume~Dubuisson Duplessis, Vincent Letard, Anne{-}Laure Ligozat, and Sophie
  Rosset.
\newblock Purely corpus-based automatic conversation authoring.
\newblock In {\em Proceedings of the Tenth International Conference on Language
  Resources and Evaluation {LREC} 2016, Portoro{\v{z}}, Slovenia, May 23-28,
  2016.}, 2016.

\bibitem{elson2012dramabank}
David Elson.
\newblock Dramabank: Annotating agency in narrative discourse.
\newblock In {\em LREC}, pages 2813--2819, 2012.

\bibitem{NPS}
Eric~N. Forsythand and Craig~H. Martell.
\newblock Lexical and discourse analysis of online chat dialog.
\newblock In {\em Proceedings of the International Conference on Semantic
  Computing}, ICSC '07, pages 19--26, Washington, DC, USA, 2007. IEEE Computer
  Society.

\bibitem{gilmartin17}
Emer Gilmartin, Benjamin~R. Cowan, Carl Vogel, and Nick Campbell.
\newblock Chunks in multiparty conversation - building blocks for extended
  social talk.
\newblock In {\em IWSDS '17}, 2017.

\bibitem{GS86}
Barbara~J. Grosz and Candace~L. Sidner.
\newblock Attention, intentions and the structure of discourse.
\newblock {\em Computational Linguistics}, 12:175--204, 1986.

\bibitem{guzzoni2007active}
Didier Guzzoni, Adam Cheyer, and Charles Baur.
\newblock Active, a platform for building intelligent operating rooms.
\newblock In {\em Surgetica 2007 Computer-Aided Medical Interventions: tools
  and applications}, pages 191--198. Sauramps M{\'e}dical, 2007.

\bibitem{hakkani2014probabilistic}
Dilek Hakkani-T{\"u}r, Asli Celikyilmaz, Larry Heck, Gokhan Tur, and Geoff
  Zweig.
\newblock Probabilistic enrichment of knowledge graph entities for relation
  detection in conversational understanding.
\newblock In {\em Proceedings of INTERSPEECH}, 2014.

\bibitem{Higashinaka14opendomain}
Ryuichiro Higashinaka, Kenji Imamura, Toyomi Meguro, Chiaki Miyazaki, Nozomi
  Kobayashi, Hiroaki Sugiyama, Toru Hirano, Toshiro Makino, and Yoshihiro
  Matsuo.
\newblock Towards an open-domain conversational system fully based on natural
  language processing.
\newblock In {\em Proceedings of COLING 2014, the 25th International Conference
  on Computational Linguistics: Technical Papers}, pages 928--939. Dublin City
  University and Association for Computational Linguistics, 2014.

\bibitem{Hobbs78}
Jerry~R. Hobbs.
\newblock Why is discourse coherent?
\newblock Technical Report 176, SRI International, 333 Ravenswood Ave., Menlo
  Park, Ca 94025, 1978.

\bibitem{HE79}
J.R. Hobbs and D.~Evans.
\newblock Conversation as planned behaviour.
\newblock Technical Report 203, Artificial Intelligence Center, SRI
  International, Menlo Park, CA, 1979.

\bibitem{juraska2018slug2slug}
Juraj Juraska, Panagiotis Karagiannis, Kevin~K. Bowden, and Marilyn~A. Walker.
\newblock A deep ensemble model with slot alignment for sequence-to-sequence
  natural language generation.
\newblock {\em NAACL}, 2018.

\bibitem{kamm1997evaluating}
Candace Kamm, Shrikanth Narayanan, Dawn Dutton, and Russell Ritenour.
\newblock Evaluating spoken dialog systems for telecommunication services.
\newblock In {\em Fifth European Conference on Speech Communication and
  Technology}, 1997.

\bibitem{krause2017edina}
Ben Krause, Marco Damonte, Mihai Dobre, Daniel Duma, Joachim Fainberg, Federico
  Fancellu, Emmanuel Kahembwe, Jianpeng Cheng, and Bonnie Webber.
\newblock Edina: Building an open domain socialbot with self-dialogues.
\newblock {\em arXiv preprint arXiv:1709.09816}, 2017.

\bibitem{krause2017redundancy}
Sebastian Krause, Mikhail Kozhevnikov, Eric Malmi, and Daniele Pighin.
\newblock Redundancy localization for the conversationalization of unstructured
  responses.
\newblock In {\em Proceedings of the 18th Annual SIGdial Meeting on Discourse
  and Dialogue}, pages 115--126, 2017.

\bibitem{LisonTK18opensubtitles}
Pierre Lison, J{\"{o}}rg Tiedemann, and Milen Kouylekov.
\newblock Opensubtitles2018: Statistical rescoring of sentence alignments in
  large, noisy parallel corpora.
\newblock In {\em Proceedings of the Eleventh International Conference on
  Language Resources and Evaluation, {LREC} 2018, Miyazaki, Japan, May 7-12,
  2018.}, 2018.

\bibitem{Litman85}
Diane Litman.
\newblock Plan recognition and discourse analysis: An integrated approach for
  understanding dialogues.
\newblock Technical Report 170, University of Rochester, 1985.

\bibitem{Louisetal10}
Annie Louis, Aravind Joshi, Rashmi Prasad, and Ani Nenkova.
\newblock Using entity features to classify implicit relations.
\newblock In {\em Proc. of the 11th Annual SIGdial Meeting on Discourse and
  Dialogue}, Tokyo, Japan, 2010.

\bibitem{Personabank}
Stephanie~M. Lukin, Kevin Bowden, Casey Barackman, and Marilyn~A. Walker.
\newblock A corpus of personal narratives and their story intention graphs.
\newblock In {\em In Proceedings of the 10th International Conference on
  Language Resources and Evaluation (LREC)}, 2016.

\bibitem{mann88}
William~C. Mann and Sandra~A. Thompson.
\newblock Rhetorical structure theory. {T}oward a functional theory of text
  organization.
\newblock {\em Text}, 8(3):243--281, 1988.

\bibitem{stanford14}
Christopher Manning, Mihai Surdeanu, John Bauer, Jenny Finkel, Steven Bethard,
  and David McClosky.
\newblock The stanford corenlp natural language processing toolkit.
\newblock In {\em Proceedings of 52nd Annual Meeting of the Association for
  Computational Linguistics: System Demonstrations}, pages 55--60. Association
  for Computational Linguistics, 2014.

\bibitem{MarcuEchihabi02}
D.~Marcu and A.~Echihabi.
\newblock An unsupervised approach to recognizing discourse relations.
\newblock In {\em Proc. of the 40th Annual Meeting on Association for
  Computational Linguistics}, pages 368--375, 2002.

\bibitem{marcu96}
Daniel Marcu.
\newblock Building up rhetorical structure trees.
\newblock In {\em Proc. of {AAAI}/{IAAI} 1996}, volume~2, pages 1069--1074,
  1996.

\bibitem{Nio2014}
Lasguido Nio, Sakriani Sakti, Graham Neubig, Tomoki Toda, Mirna Adriani, and
  Satoshi Nakamura.
\newblock {\em Developing Non-goal Dialog System Based on Examples of Drama
  Television}, pages 355--361.
\newblock Springer New York, New York, NY, 2014.

\bibitem{Prasadetal10}
R.~Prasad, A.~Joshi, and B.~Webber.
\newblock Realization of discourse relations by other means: alternative
  lexicalizations.
\newblock In {\em Proc. of the 23rd International Conference on Computational
  Linguistics: Posters}, pages 1023--1031. Association for Computational
  Linguistics, 2010.

\bibitem{prasad-etal-lrec08}
Rashmi Prasad, Nikhil Dinesh, Alan Lee, Eleni Miltsakaki, Livio Robaldo,
  Aravind Joshi, and Bonnie Webber.
\newblock The {Penn Discourse TreeBank 2.0.}
\newblock In {\em Proc. of 6th International Conference on Language Resources
  and Evaluation (LREC 2008)}, 2008.

\bibitem{shah2016interactive}
Pararth Shah, Dilek Hakkani-T{\"u}r, and Larry Heck.
\newblock Interactive reinforcement learning for task-oriented dialogue
  management.
\newblock In {\em NIPS 2016 Deep Learning for Action and Interaction Workshop},
  2016.

\bibitem{shah2018bootstrapping}
Pararth Shah, Dilek Hakkani-Tur, Bing Liu, and Gokhan Tur.
\newblock Bootstrapping a neural conversational agent with dialogue self-play,
  crowdsourcing and on-line reinforcement learning.
\newblock In {\em Proceedings of the 2018 Conference of the North American
  Chapter of the Association for Computational Linguistics: Human Language
  Technologies, Volume 3 (Industry Papers)}, volume~3, pages 41--51, 2018.

\bibitem{Tonellietal10}
S.~Tonelli, G.~Riccardi, R.~Prasad, and A.~Joshi.
\newblock Annotation of discourse relations for conversational spoken dialogs.
\newblock In {\em Proc. of the Seventh International Conference on Language
  Resources and Evaluation (LREC 2010), Valletta, Malta}, pages 2084--2090,
  2010.

\bibitem{vaswani2017attention}
Ashish Vaswani, Noam Shazeer, Niki Parmar, Jakob Uszkoreit, Llion Jones,
  Aidan~N Gomez, {\L}ukasz Kaiser, and Illia Polosukhin.
\newblock Attention is all you need.
\newblock In {\em Advances in Neural Information Processing Systems}, pages
  5998--6008, 2017.

\bibitem{Walkeretal97}
M.~A. Walker, D.~Litman, C.~A. Kamm, and A.~Abella.
\newblock {PARADISE}: A general framework for evaluating spoken dialogue
  agents.
\newblock In {\em Proc. of the 35th Annual Meeting of the Association for
  Computational Linguistics, ACL/EACL 97}, pages 271--280, 1997.

\bibitem{Weizenbaum1966ELIZA}
Joseph Weizenbaum.
\newblock Eliza - a computer program for the study of natural language
  communication between man and machine.
\newblock {\em Commun. ACM}, 9(1):36--45, January 1966.

\end{thebibliography}

\end{document}